\documentclass[11pt]{article}       
\usepackage{geometry}               
\title{On Task-Relevant Loss Functions in Meta-Reinforcement Learning and Online LQR\thanks{This work was supported in part by the National Research Foundation of Korea funded by MSIT(2020R1C1C1009766), and the Information and Communications Technology Planning and Evaluation (IITP) grant funded by MSIT(2022-0-00124, 2022-0-00480).}}

\usepackage{times}

\usepackage{mathrsfs}
\usepackage{mathtools}
\usepackage{algorithm}
\usepackage{algorithmic}
\usepackage{bm}

\usepackage{graphicx}
\usepackage{caption}
\usepackage{subcaption}
\usepackage{enumerate}
\usepackage{hyperref}
\usepackage{upgreek}
\usepackage{todonotes}
\usepackage{cite}

\usepackage[skins, theorems]{tcolorbox}
\tcbset{highlight math style={enhanced,
  colframe=black,colback=white,arc=0pt,boxrule=1pt}}

\newcommand{\enc}{\bm{\mathrm{enc}}}
\newcommand{\latent}{\bm{\mathrm{z}}}

\usepackage{amsmath}
\usepackage{amsthm}
\usepackage{amssymb}
\newtheorem{theorem}{Theorem}

\DeclareMathOperator*{\argmax}{arg\,max}
\DeclareMathOperator*{\argmin}{arg\,min}

\author{Jaeuk Shin
\and
Giho Kim
\and
Howon Lee
\and
Joonho Han
\and
 Insoon Yang\thanks{J. Shin, G. Kim, J. Han, and I. Yang are with the Department of Electrical and Computer Engineering, Automation and Systems Research Institute,  Seoul National University, Seoul 08826, Korea, 
        {\tt\small \{sju5379, chicioue512, snowhan1021, insoonyang\}@snu.ac.kr}. H. Lee is with Interdisciplinary Program in Artificial Intelligence, ASRI, Seoul National University, Seoul 08826, Korea, {\tt\small ho1dol2@snu.ac.kr}.}  
}

\date{}

\begin{document}

\maketitle

\pagestyle{myheadings}
\thispagestyle{plain}

\begin{abstract}%
Designing a competent meta-reinforcement learning (meta-RL) algorithm in terms of data usage remains a central challenge to be tackled for its successful real-world applications. In this paper, we propose a sample-efficient meta-RL algorithm that learns a model of the system or environment at hand in a task-directed manner. As opposed to the standard model-based approaches to meta-RL, our method exploits the value information in order to rapidly capture the decision-critical part of the environment. The key component of our method is the loss function for learning the task inference module and the system model that systematically couples the model discrepancy and the value estimate, thereby facilitating the learning of the policy and the task inference module with a significantly smaller amount of data compared to the existing meta-RL algorithms. The idea is also extended to a non-meta-RL setting, namely an online linear quadratic regulator (LQR) problem, where our method can be simplified to reveal the essence of the strategy. The proposed method is evaluated in high-dimensional robotic control and online LQR problems, empirically verifying its effectiveness in extracting information indispensable for solving the tasks from observations in a sample efficient manner.
\end{abstract}


\section{Introduction}
\textit{Meta-reinforcement learning (meta-RL)} has become widely recognized and accepted in the domain of decision-making, particularly for its capability to systematically adapt to environmental changes. The main strategy of meta-RL involves devising a mechanism capable of identifying the task-specific features from data and utilizing them to generate actions tailored to the task. Unfortunately, even though the field has elicited considerable research interest in the development of a task inference module that rapidly adapts to novel tasks, a critical challenge persists: the reduction of the severe sample complexity associated with meta-training, i.e., learning such a module along with the policy from data. This inefficiency is notably revealed in popular \textit{model-free} meta-RL algorithms, including gradient-based methods~\cite{finn2017model,gupta2018meta} and off-policy context-based methods~\cite{rakelly2019efficient}, thereby impeding their application to intricate real-world decision-making scenarios.

The standard RL community has produced  a series of studies investigating the statistical advantages of model-based RL methods over model-free RL methods in either  theoretical~\cite{tu2019gap,sun2019model} or experimental~\cite{pong2018temporal} contexts. The core message from these works that the model-based methods often enjoy better sample efficiency than the model-free methods has triggered the rapid development of \textit{model-based meta-RL} methods~\cite{nagabandi2018learning,saemundsson2018meta,galashov2019meta,wang2022model}, where the model-based techniques are leveraged to alleviate the data dependency of the meta-RL methods. In~\cite{nagabandi2018learning}, a neural network model is adapted online through gradient-based updates and deployed in a model predictive control (MPC) controller. However, the gradient estimates are often noisy, especially when only a limited amount set of samples are used to construct the estimator, and the model parameters are only locally updated when combined with techniques such as trust region policy optimization~\cite{schulman2015trust}. Accordingly, it is unsurprising that such a method requires a large amount of data when meta-testing, which is undesirable for real-world applications that must rapidly adapt to environmental changes. In~\cite{saemundsson2018meta}, the Gaussian process (GP) is used to model the unknown system dynamics, and to infer the model online, a context-based approach is incorporated into the method by introducing latent variables that represent task information. However, learning and inferring the model are done independently of the quality of the resulting policies. In~\cite{shin2022infusing}, MPC controllers with local GP models are employed in an event-based manner for generating high-quality action samples in order to derive an efficienct the meta-RL algorithm, but again the models are learned without the consideration of the value information. These indicate that the data efficiency of the approach can be improved by focusing on accurately learning and identifying the \textit{task-relevant} part of the system dynamics while neglecting the remaining part.

In model-free methods, there have been numerous attempts to define the notion of \textit{task-relevant} information and to develop the methods that efficiently capture and exploit such information. The intuition behind these approaches is that while each task is characterized by its dynamics (or its reward function occasionally), the perfect description of the dynamics may not be essential for optimal decision-making. For instance, \text{PEARL}~\cite{rakelly2019efficient} trains its task inference module to learn the $Q$-function that exhibits a small temporal difference (TD) error, in which case the task inference directly corresponds to identifying the optimal policy from the data. In~\cite{fu2021towards}, the noisy nature of TD errors is highlighted as a main issue of performing such task inference, and a novel task recognition method based on contrastive learning is proposed. Finally, \cite{liu2021decoupling} presents a mutual information minimization method that can neglect any task-irrelevant information for detecting the optimal value function of each task. However, to the best of our knowledge, efficiently extracting task-relevant information has  been sparsely studied in a model-based context.

To faithfully learn the task-relevant portion of the task dynamics and accelerate the meta-learning procedure, we focus on the suboptimality bound of the policy derived from an approximate model, a metric that essentially quantifies the divergence in value prediction. This leads us to introduce \textit{\textbf{t}ask-\textbf{r}elevant \textbf{m}eta-\textbf{r}einforcement \textbf{l}earning (TRMRL)}, a novel model-based method that systematically exploits reward information when learning task-adaptive models. Our method proceeds by inferring the environmental model of a task from data, where we train the inference module to yield a model that precisely estimates the nominal values. The underlying idea for this design is that the model does not have to exhaustively capture the full dynamics of the system; rather, only its capability of correctly predicting state values does matters for obtaining a near-optimal policy. Such an effort to recognize task-centric information promotes sample efficiency, thereby advancing meta-RL  toward applicability in problems with limited opportunities to collect data.

Our contributions are summarized as follows:
\begin{itemize}
\item We propose a theoretically principled loss function for learning task-directed models of unknown system components, namely the system dynamics and the reward function. The loss function is carefully designed through the analysis of the policy's suboptimality bound that involves the value estimation capability of the learned model.

\item Using the loss function, we design a novel model-based meta-RL method that selectively learns the task-relevant models of a task, The method is highly efficient in that it uses a notably smaller amount of data compared to the existing meta-RL methods to present a comparable performance. This is achieved by training both the task inference module and the model of the system dynamics and the reward function using the proposed loss function, which enables the task inference module to recognize a policy that is adequate for a new task. 

\item We present a novel algorithm called \texttt{\textbf{TR-SGD}} for online linear quadratic regulator (LQR) problems. This algorithm is also built upon the task-relevant loss function.

\item Our meta-RL method is experimentally evaluated in a complex robotic control problem, where the physical properties of the robot and the environment vary across the tasks. 
The result of our experiments show that 
our method outperforms other state-of-the-art  off-policy meta-RL methods in terms of sample efficiency and average return.  We also confirm the efficiency of \texttt{\textbf{TR-SGD}} in a high-dimensional online LQR problem.
\end{itemize}

Throughout the paper, we let $\Delta(\mathcal{X})$ denote the space of all probability measures defined over a set $\mathcal{X}$. The notation $\delta_x$ is used to denote the Dirac measure on $x \in \mathcal{X}$. For $P \in \Delta(\mathcal{X})$ and $f \in L^1(\mathcal{X}, P)$,   $\mathbb{E}_{x \sim P} [f(x)]$ denotes the expectation of $f$ with respect to $P$.

\section{Background and Motivation}

Recent developments of data-driven sequential decision-making methods necessitate a principled method toward the efficient control of complex real-world systems, most of which involve systems characterized by environmental variability. In practice, such changes are hardly anticipated by decision-makers \textit{a priori}, which forces the decision-making agents to deduce these changes from experimental data and adapt their policies accordingly. Consequently, modeling these problems as single-task problems is unreliable, primarily because using a policy calibrated for a single system model may lead to catastrophic outcomes when it faces environmental changes~\cite{nagabandi2018learning}. To systematically address the \textit{multi-task} challenge of real-world scenarios, meta-RL has been actively studied~\cite{wang2016learning,duan2016rl,finn2017model,nagabandi2018learning,beck2023survey}. Mathematically, meta-RL considers a collection $\mathcal{M}$ of \textit{Markov decision processes (MDPs)}~\cite{puterman2014markov}, or a collection of \textit{tasks}, equivalently, and a \textit{task distribution} $\mathscr{T}$ over $\mathcal{M}$. Each task consists of data $M = \langle\mathcal{S}, \mathcal{A}, T^M, T_0, R^M, \gamma\rangle \in \mathcal{M}$ where $\mathcal{S}$ is a state space, $\mathcal{A}$ is an action space, $T^M: \mathcal{S} \times \mathcal{A} \to \Delta(\mathcal{S})$ is a stochastic transition kernel that describes the dynamics of the system, $T_0 \in \Delta(\mathcal{S})$ is an initial state distribution, $R^M : \mathcal{S} \times \mathcal{A} \to \mathbb{R}$ is a reward function, and $\gamma \in [0, 1)$ is a discount factor. In each episode, a task $M = \langle T^M, R^M \rangle$ is randomly drawn from $\mathscr{T}$, where a decision-maker receives state and reward measurements from $M$. Specifically, a \textit{policy} $\pi$ determines an action $a_k \sim \pi(\cdot | h_k)$ to execute based on the history $h_k \coloneqq (s_0, a_0, r_0, s_1, \ldots, s_{k-1}, a_{k-1}, r_{k-1}, s_k)$ of the states, actions, and rewards up to $k$.
 Then, the goal of meta-RL is to solve the following policy optimization problem:
\begin{equation}\label{eq:meta-mdp-objective}
\sup_{\pi} \mathop{\mathbb{E}}_{M \sim \mathscr{T}} \mathbb{E}^{T_0, T^M, \pi} \left[ \sum_{k =0}^{\infty} \gamma^k R^M(s_k, a_k) \right],
\end{equation}
where the inner expectation is computed over the following state and action process $s_0 \sim T_0,\; a_{k} \sim \pi(\cdot | h_k),\;  s_{k+1} \sim T(\cdot | s_k, a_k),\;  k = 0, 1, \ldots$.
Therefore, the policy needs to act adaptively to an unknown MDP $M$ as identifying essential information about $M$ from the collected history. The problem~\eqref{eq:meta-mdp-objective} is rigorously stated as a standard RL problem called \textit{Bayes-adaptive MDP (BAMDP)}~\cite{duff2002optimal} by introducing a notion of \textit{belief state} that represents the probabilistic estimate of the model $T^M$ and $R^M$ of a given task. Formally, a belief state $b_k$ given a history $h_k$ is defined as a posterior distribution $b_k = b(\cdot | h_k)$ over $\mathcal{M}$ determined via Bayes' theorem:
\begin{equation}\label{eq:belief-update}
b( M | h_k) \propto \mathscr{T}(M) \prod_{i = 0}^{k-1}  T^M(s_{i+1} | s_i, a_i) \mathbb{I}_{R^M(s_i, a_i)}(r_i),
\end{equation}
and an optimal policy $\pi^\star(s_k, b_k)$ takes $s_k$ and $b_k$ as its inputs instead of the entire history $h_k$ to yield an action $a_k$. 
Unfortunately, solving BAMDPs suffers from the curse of dimensionality, as \textit{(i)} performing the update~\eqref{eq:belief-update} is generally known as intractable if the task distribution is high-dimensional even when the exact information about each $P^M$ and $R^M$ is given, and \textit{(ii)}, computing the near-optimal value function requires excessive computation when the belief state is high-dimensional. Instead, the posterior distribution in~\eqref{eq:belief-update} is often approximated as a distribution over a low-dimensional space called \textit{latent space} where each latent vector corresponds to a single task. Such a method is called \textit{context-based meta-RL}~\cite{rakelly2019efficient,zintgraf2019varibad,zhang2020learning}. Specifically, in context-based meta-RL, each MDP $M$ is assumed to be embedded into a low-dimensional \textit{context variable} $\textbf{z} \in Z \coloneqq \mathbb{R}^d$. This is done by a task inference module called a \textit{context encoder} $\enc$ which consumes data $\mathcal{D} = \{(s_i, a_i, r_i, s_{i+1}) \}$ from $M$ and generates a latent distribution, denoted as
\[
\mathcal{P}(d\latent) = \enc(\mathcal{D}) \in \Delta(\mathbb{R}^d), 
\]
which can be understood as a proxy of the posterior distribution~\eqref{eq:belief-update} inferred from $\mathcal{D}$ but in the latent space $Z$ instead of the space of MDPs $\mathcal{M}$. Finally, a context variable $\latent$ is sampled from $\mathcal{P}(d\latent)$ which is then used as an input of a policy $\pi_{\latent}(s)$, enabling it to perform task-adaptive actions, or models $T_{\latent}(s'|s, a)$ and $R_{\latent}(s, a)$. The encoder may be jointly trained with a policy (and the corresponding value function) in a way that the resulting policy $\pi_{\latent}(s)$ is near-optimal for $M$~\cite{rakelly2019efficient} or the log-likelihood of sample trajectories is maximized under $T_{\latent}(s'|s, a)$ and $R_{\latent}(s, a)$ via approximate inference~\cite{zintgraf2019varibad}. These methods are more advantageous than other meta-RL methods that go through gradient-based searches in large parameter spaces for policy adaptation~\cite{finn2017model,gupta2018meta,nagabandi2018learning} in terms of data utilization since the task inference can be done with a relatively small amount of data and computational resources. Nevertheless, they suffer from sample inefficiency when it comes to meta-training the policies and the task inference modules. For instance, VariBAD~\cite{zintgraf2019varibad} aims to attain a near-optimal solution of~\eqref{eq:meta-mdp-objective} through the direct modeling of~\eqref{eq:belief-update}. The method is an on-policy algorithm, which leads to poor sample complexity. Despite being an off-policy method, PEARL~\cite{rakelly2019efficient} is a model-free algorithm; thus, there is room to  further improve the efficient exploitation of the collected data.

\section{Task-Relevant Loss Functions}
We begin with presenting a general form of the task-relevant loss functions, which will  be tailored to meta-RL and online LQR settings. Our loss function is inspired by the following theorem, which can be viewed as a generalization of Theorem 7 of~\cite{pires2016policy}:
\begin{theorem}\label{thm:main}
Suppose that $\{ M_{\latent} = \langle T_{\latent}, R_{\latent}\rangle \}_{\latent\in Z}$ is a collection of tasks parameterized by $\latent \in Z$, and $\{ V_{\latent} : \mathcal{S} \to \mathbb{R} \}_{\latent \in Z}$ is a collection of value functions satisfying $\Vert V^\star_{M_{\latent}} - V_{\latent} \Vert_\infty \leq \varepsilon$ for each $M_{\latent}$, 
where $V^\star_{M_{\latent}}$ denotes the optimal value function for $M_{\latent}$. Let $V^\pi_M$ be the value function of $\pi$ for $M$, and $(T V) (s, a) \coloneqq \mathbb{E}_{s'\sim T(\cdot|s, a)}\left[ V(s') \right]$.
If $\pi_{\latent}$ is greedily constructed from  $M_{\latent}$ and $V_{\latent}$, i.e.,
\begin{equation*}
\pi_{\latent}(s) \in \argmax_a  \Big ( R_{\latent}(s, a) + \gamma \mathbb{E}_{s' \sim T_{\latent}(\cdot | s, a)} \left[ V_{\latent}(s')\right] \Big), \quad s \in \mathcal{S},
\end{equation*}
then for any task $M = \langle T, R \rangle \in \mathcal{M}$ and $\latent \in Z$, the following inequality holds:
\begin{equation}\label{eq:approx-ineq}
\left\Vert V_M^\star - V_M^{\pi_{\latent}} \right\Vert_\infty \leq \underbrace{\frac{2}{1 - \gamma} \left\Vert (R - R_{\latent}) + \gamma\left(T - T_{\latent} \right) V_{\latent} \right\Vert_\infty}_{\textbf{task inference error}} + \underbrace{\frac{(4 + 2\gamma(1 - \gamma))\varepsilon }{(1 - \gamma)^2}}_{\textbf{planning error}}.
\end{equation}
\end{theorem}
The proof of Theorem~\ref{thm:main} can be found in Appendix~\ref{app:proof}. Theorem~\ref{thm:main} may be regarded as echoing the prevalent assertion that when the model used for policy planning precisely reflects the actual system (task inference) and the policy is near-optimal for this model (planning), then the policy is near-optimal for the true task. However, the error bound~\eqref{eq:approx-ineq} uncovers a more nuanced relationship between the quality of the learned model and the suboptimality of the policy: Quantifying a model mismatch without considering the value might be conservative. Intuitively, Theorem~\ref{thm:main} suggests that for seeking a near-optimal policy for a specific task $M$, it is sufficient for the encoder to generate $\latent$ whose associated model $\langle T_{\latent}, R_{\latent} \rangle$ aligns accurately with the value across states and actions rather than precisely replicating the transitions. Indeed, correctly guessing the value of $R + \gamma T V_{\latent}$ hinges on matching the expectation of $V_{\latent}$ over $T(\cdot|s, a)$, which is markedly less demanding than precisely estimating the entire transition probability distribution.
Accordingly, using~\eqref{eq:approx-ineq} for learning enables us to rapidly obtain an approximate model that is sufficient for generating a high-performance policy even if it is not capable of exactly matching the state transitions. Furthermore, as pointed out in \cite{pires2016policy}, the task inference error in~\eqref{eq:approx-ineq} does not depend on the optimal value function, as the right-hand side of~\eqref{eq:approx-ineq} measures the difference between the nominal model and the true model in terms of their estimated value $V_{\latent}$. Hence, the quantity $ \left(R - R_{\latent}\right) + \gamma\left(T - T_{\latent} \right) V_{\latent}$ can be explicitly constructed and used for learning in practice.

\subsection{Meta-RL}
A major challenge of applying the widely known meta-RL methods to real-world problems is their notorious sample inefficiency; for a solution to be feasible when collecting a large amount of data is prohibitive, it must exhibit significantly higher statistical complexity than the existing methods. Although there have been attempts to leverage techniques from model-based reinforcement learning to resolve the problem~\cite{perez2018efficient,nagabandi2018learning,saemundsson2018meta,belkhale2021model}, the central question of how to exploit the cost information for learning the models has not yet been appropriately addressed. To enhance the sample efficiency of model-based methods for meta-RL problems, we exploit the result of Theorem~\ref{thm:main} to propose a novel method called \textit{task-relevant meta-reinforcement learning (TRMRL)}, a  model-based meta-RL algorithm that effectively extracts task-relevant information from data.

As a model-based method, TRMRL maintains the models $T_{\theta, \latent}(s'|s, a)$ and $R_{\theta, \latent}(s, a)$ of the MDP dynamics and the reward function, respectively, where $\theta$ represents learnable parameters, and $\textbf{z}$ is a context variable that is inferred from a set of transitions. Specifically, when a novel task $M$ is drawn from $\mathscr{T}$ and $\mathcal{D} = \{ (s_i, a_i, r_i, s_{i+1}) \}_{i = 1}^N$ is observed, then the output of the context encoder $\mathcal{P}(d\latent) = \enc_\phi(\mathcal{D})$ induces a probability distribution over the models
\begin{equation*}
M_{\theta, \textbf{z}} \coloneqq \langle T_{\theta, \latent}(s'|s, a), R_{\theta, \latent}(s, a) \rangle, \quad \latent \sim \mathcal{P}(d\latent) = \enc_\phi(\mathcal{D}),
\end{equation*}
which represents the probabilistic estimate of $M$. For each $\latent$, a near-optimal policy of an MDP $M_{\theta, \latent}$ may be obtained by applying planning techniques such as policy iteration.
Then, one of the most straightforward ways of updating our model is to maximize the log-likelihood of another dataset $\mathcal{D}'$ from $M$. For instance, the loss function for the transition model may be constructed as
\begin{equation}\label{eq:model-loss-naive}
\mathcal{L}(\theta, \phi) \coloneqq \frac{1}{|\mathcal{D}'|} \sum_{(s, a, s')\in \mathcal{D}'}  \log \mathbb{E}_{\textbf{z}\sim \enc_\phi(\mathcal{D})} T_{\theta, \latent}(s'| s, a),
\end{equation}
which can be used for updating both $T_\theta$ and $\enc_\phi$.
A limitation of using~\eqref{eq:model-loss-naive} is that it is completely independent of the structure of the underlying decision-making problem. To address this issue, we refine~\eqref{eq:model-loss-naive} to reflect the value information of the problem. Specifically, our method is based on~\eqref{eq:approx-ineq} whose right hand side is directly taken as the loss function for both the encoder $\enc_{\phi}$ and the model $T_{\theta}$. It is logical to expect such an application, since Theorem~\ref{thm:main} provides a clear answer to which $\latent$ the encoder has to generate to obtain a task-adaptive policy. Indeed, the encoder can be trained so that, given $\mathcal{D}$, it selects $\latent$ which minimizes the task inference error.
This motivates using the following type of loss functions to train both the model and the encoder:
\begin{equation}\label{eq:encoder-loss}
\tcbhighmath[left=2pt,right=2pt,top=1pt,bottom=1pt]{
\mathcal{L}(\theta, \phi) \coloneqq \mathop{\mathbb{E}}_{\substack{M \sim \mathscr{T}_{\text{train}} \\ \mathcal{D}, \mathcal{D}' \sim M}} 
 \left\{ \mathop{\mathbb{E}}_{\substack{\textbf{z} \sim \enc_{\phi}(\mathcal{D}) \\ (s, a, r, s') \sim \mathcal{D}'}}\left( r - R_{\theta, \latent}(s, a) + \gamma V_{\psi, \latent}(s') - \gamma \mathop{\mathbb{E}}_{ T_{\theta, \latent}(s' | s, a)} V_{\psi, \latent}(s') \right)^2 \right\}
 }
\end{equation}
where $\mathscr{T}_{\text{train}}$ denotes the training task distribution, $\mathcal{D}$ is for task inference, and $\mathcal{D}'$ is for evaluating $\left(R - R_{\latent}\right) + \gamma\left(T - T_{\latent} \right) V_{\latent}$ over a subset of $\mathcal{S} \times \mathcal{A}$.
Note that the terms $r$ and $V_{\psi, \latent}(s')$ in~\eqref{eq:encoder-loss} serves as approximations for $R(s, a)$ and $\mathbb{E}_{T(s'|s, a)} [V_{\psi, \latent}(s')]$, respectively. A similar observation has been exploited to design a loss function for model learning~\cite{farahmand2018iterative} in the standard RL setting, where only the  model of the system dynamics is uncertain.

The outline of TRMRL is given in Algorithm~\ref{alg:trmrl}. The algorithm consists of two parts: a data collection stage and a training stage. In the data collection stage, a set of training tasks is sampled, and the latent variable $\latent_0$ is drawn from the prior distribution $\mathcal{P}_0(d\latent)$. The prior distribution may be thought of as representing the entire task distribution, as no information about a given task is assigned. However, after each trajectory $\tau^{j}$ is observed by executing the policy $\pi_{\psi', \latent_j}(a|s)$, the posterior distribution is updated to $\mathcal{P}_{j+1}(d\latent)$ by the encoder $\enc_{\phi}$ and a new latent vector $\latent_{j+1}$ is given as a new input to the policy $\pi_{\psi', \latent}$.
In the training stage, the policy $\pi_{\latent}$ and the approximate value function $V_{\latent}$ for each task $M_{\latent}$ are learned through planning (lines 13--14). In particular, training the policy and the value functions may be done using synthetic data generated from the model $T_{\theta, \latent}$ and $R_{\theta, \latent}$, thereby taking advantage of the sample efficiency of model-based methods.
Indeed, we apply a simple Dyna-style method~\cite{sutton1991dyna,munos2008finite} for model-based learning and planning.\footnote{There are possible alternatives, such as using model predictive control~\cite{lowrey2018plan}.} Using this method, we learn approximate value functions $V_{\psi, \latent}(s)$ and the policies $\pi_{\psi', \latent}(s)$ of the entropy-regularized MDPs~\cite{haarnoja2018soft,geist2019theory}. The reason for planning on the entropy-regularized MDPs is that the learned policies are capable of exploring the state and action spaces and therefore collecting data sufficiently diverse for task identification. Finally, the model $\langle T_\theta, R_\theta \rangle$ and the encoder $\enc_\phi$ are trained using the task-relevant loss function~\eqref{eq:encoder-loss} (lines 15--16), which makes it possible to pre-emptively learn value-critical parts of the dynamics.
\begin{algorithm}[t!]
\caption{Task-relevant meta-reinforcement learning (TRMRL)}
\label{alg:trmrl}
\begin{algorithmic}[1]
	\STATE \textbf{Input}: task batch size $N_{\text{task}} > 0$, training task distribution $\mathscr{T}_{\text{train}}$, $N_\text{episode}$, prior latent distribution $\mathcal{P}_0 = \mathcal{N}(0, I)$, replay buffer $\mathcal{B}_i$ for each training task $M_i$
	\FOR {$k = 1,2, \ldots $}
              \STATE {\color{violet}\texttt{// data collection}}
              \FOR {$i = 1,\ldots, N_{\text{task}}$}
	       \STATE Sample a training task $M_i \sim \mathscr{T}_{\text{train}}$ and randomly drawn a latent vector $\latent_0 \sim \mathcal{P}_0(d\latent)$;
                \FOR{episode $j=0,\ldots, N_{\text{episode}}-1$}
                    \STATE Collect a trajectory $\tau^j$ by executing $\pi_{\psi', \latent}(a|s_j)$;
                    \STATE Add $\tau^j$ to the replay buffer $\mathcal{B}_i$;
                    \STATE Infer the posterior distribution $\mathcal{P}_{j+1}(d\latent) \coloneqq \enc_\phi(\tau^0, \ldots, \tau^j)$ and sample $\latent_{j+1} \sim \mathcal{P}_{j+1}(d\latent)$;
                \ENDFOR
             
              \ENDFOR
              \STATE {\color{violet}\texttt{// training}}
              \STATE Randomly sample $\textbf{z}\sim \mathcal{P}_0$ \& randomly generate state-action pairs $B = \{ (s^\ell, a^\ell) \}$;
              \STATE Learn $\pi_{\psi', \latent}$ and $V_{\psi, \latent}$ for the model $M_{\theta, \latent}$ using $B$;
	       \STATE Construct task inference loss~\eqref{eq:encoder-loss} by sampling $\mathcal{D}$ \& $\mathcal{D}'$ from $\mathcal{B}_i$'s;
              \STATE Perform the gradient descent updates to learn $\enc_\phi$ and $T_\theta$:
              \begin{align*}
              \theta &\longleftarrow \theta - \alpha_{\theta} \nabla_{\theta}\mathcal{L}(\theta, \phi), \qquad \phi \longleftarrow \phi - \alpha_{\phi} \nabla_{\phi}\mathcal{L}(\theta, \phi).
              \end{align*}
	\ENDFOR
\end{algorithmic}
\end{algorithm}

\subsection{Online LQR}
The strategy of exploiting task-directed inequality~\eqref{eq:approx-ineq} for learning the system dynamics not only promotes the faster learning of meta-RL methods but  also  may be applied to improve general model-based control methods. As a popular example, we consider  an \textit{online linear quadratic regulator (LQR)} with unknown system parameters, where the planning error in~\eqref{eq:approx-ineq} vanishes. The system dynamics and the reward function of LQR problems are given as
\begin{equation*}
x_{k+1} = Ax_k + Bu_k, \quad r(x_k, u_k) \coloneqq -(x_k^\top \mathbf{Q} x_k + u_k^\top \mathbf{R} u_k),\quad k = 0, 1, \ldots,
\end{equation*}
 where $(x_k, u_k) \in \mathbb{R}^n \times \mathbb{R}^m$, and $\mathbf{Q} = \mathbf{Q}^\top \succeq 0$, $\mathbf{R} = \mathbf{R}^\top \succ 0$ are known. When $(\sqrt{\gamma}A, \sqrt{\gamma}B)$ is stabilizable and $(\sqrt{\gamma}A, \mathbf{Q}^{1/2})$ is observable, the optimal value function $V(x) \coloneqq \sup_{\pi} \mathbb{E}[\sum_{k=0}^\infty \gamma^k r(x_k, u_k)  | x_0 = x ]$
and the optimal policy are given by
 \begin{equation}\label{eq:lq-optimal}
V(x) = x^\top P^\star x, \quad \pi(x) = \gamma(\mathbf{R} + \gamma B^\top P B)^{-1} B^\top PA x,
 \end{equation}
 where $P^\star = {P^\star}^\top \succ 0$ is the solution of the following \textit{discrete-time algebraic Riccati equation (DARE)}:
 \begin{equation*}
P = \mathbf{Q} + \gamma A^\top P A - \gamma^2 A^\top P B (\mathbf{R} + \gamma B^\top P B)^{-1}B^\top P A.
 \end{equation*}
The model parameters $\Theta^\star \coloneqq (A, B)$ are assumed to be \textit{unknown}, and only the nominal parameters $\Theta^{(0)} \coloneqq (A^{(i)}, B^{(i)})$ of the true system are given. Instead, it is allowed to observe multiple trajectories
\begin{equation*}
\tau^{(i)} = \big(x^{(i)}_0, u^{(i)}_0, \ldots, x^{(i)}_N \big), \qquad i = 0, 1, \ldots
\end{equation*}
generated by applying random control inputs $u^{(i)}_k \sim \mathcal{N}(0, \sigma_u^2 I)$ starting from $x_0 = 0$.\footnote{This may be generalized to the case where a noisy certainty-equivalent controller $u_k = K x_k + \eta_k$ is used, where $\eta_k$ is white noise as in~\cite{mania2019certainty}. Indeed, such a strategy is known to achieve optimal regret even when the system is subject to a stochastic disturbance~\cite{simchowitz2020naive}.}

These trajectories may be exploited to refine our estimates about the true system parameters. One straightforward way of utilizing the data is to consider the following \textit{ordinary least-square (OLS)} loss that is completely ignorant of the value information:
\begin{equation}
J_{\text{OLS}}(\Theta| \tau) \coloneqq \frac{1}{N} \sum_{k=0}^{N-1} \left\Vert x_{k+1} - \Theta z_{k} \right\Vert_2^2 + \lambda \Vert \Theta \Vert_F^2 \quad \left(z_k \coloneqq \begin{bmatrix} x_k \\ u_k \end{bmatrix}\right)
\end{equation}
with a regularization parameter $\lambda > 0$, and update our estimates via stochastic gradient descent, which we call \textit{ordinary least-square stochastic gradient descent} (\textbf{\texttt{OLS-SGD}}):
\begin{equation}\label{eq:lq-ols-gd}
\textbf{(\texttt{OLS-SGD})}\qquad \Theta^{(i)} \longleftarrow \Theta^{(i)} - \alpha_{\text{OLS}}^{(i)} \nabla_{\Theta} J_{\text{OLS}}(\Theta^{(i)}|\tau^{(i)}), \qquad i = 1, 2, \ldots.
\end{equation}
On the other hand, a single-task version of Theorem~\ref{thm:main}, namely the case with $Z = \{\Theta \}$ and $T_{\Theta}(\cdot | z) \coloneqq T_{\Theta}(\cdot|x, u) = \delta_{\Theta z}$, suggests a entirely different form of the objective function. If $\Theta$ represents a candidate model, then both the value function $V_{\Theta}(x) = -x^\top P(\Theta) x$ and the associated greedy policy $\pi_{\Theta}$ can be computed by solving DARE with $\Theta$, without suffering from the planning error, \textit{i.e.}, $\varepsilon=0$. Furthermore, the assumption that the reward function is known corresponds to $R_\Theta = R$. As a result, the LQR version of the term $(R - R_{\Theta}) + \gamma (T - T_{\Theta})V_{\Theta}$ evaluated at $z = [x^\top \; u^\top]^\top$ is given as
\begin{equation}\label{eq:lqr-core-expression}
V_{\Theta}(x') - V_\Theta(\Theta z) = -{x'}^\top P(\Theta) {x'} + z^\top \Theta^\top P(\Theta) \Theta z,
\end{equation}
where we use the abbreviation $x' = \Theta z$ for the true next state given $x$ and $u$.
Accordingly, the task-directed loss~\eqref{eq:approx-ineq} inspires using the following loss function which evaluates the term~\eqref{eq:lqr-core-expression} along the trajectory $\tau$:
\begin{equation}\label{eq:lq-tr-obj}
\tcbhighmath[top=1pt,bottom=1pt]{
J_{\text{TR}}(\Theta|\tau, P) \coloneqq \frac{1}{N} \sum_{k = 0}^{N-1} \left\vert x_{k+1}^\top P x_{k+1} - (\Theta z_k)^\top P (\Theta z_k)  \right\vert + \lambda \Vert \Theta \Vert_F^2
}
\end{equation}
with a parameter matrix $P = P^\top \succ 0$.
Then, \textit{task-relevant stochastic gradient descent} (\textbf{\texttt{TR-SGD}}) employs~\eqref{eq:lq-tr-obj} to successively generate the parameter estimates via gradient descent:
\begin{equation}\label{eq:lq-tr-gd}
\textbf{(\texttt{TR-SGD})}\qquad\Theta^{(i)} \longleftarrow \Theta^{(i)} - \alpha_{\text{TR}}^{(i)} \nabla_{\Theta} J_{\text{TR}}(\Theta^{(i)}|\tau^{(i)}, P^{(i)}),
\end{equation}
where $P^{(i)}$ is obtained by solving DARE with data $(\sqrt{\gamma}A^{(i)}, \sqrt{\gamma}B^{(i)}, \mathbf{Q}, \mathbf{R})$.\footnote{A variant of the term of type~\eqref{eq:lq-tr-obj} frequently appears in the regret analysis of online LQR~\cite{abbasi2011regret,abeille2017thompson}. Nevertheless, to the best of our knowledge, the function~\eqref{eq:lq-tr-obj} has never been directly used as an objective function for learning the system parameters.} Once the parameter $\Theta^{(i)} = (A^{(i)}, B^{(i)})$ is obtained by using either~\eqref{eq:lq-ols-gd} or~\eqref{eq:lq-tr-gd}, a \textit{certainty-equivalent controller} $\pi^{(i)}(x) = K^{(i)} x$ is considered where the control gain matrix $K^{(i)}$ comes from~\eqref{eq:lq-optimal} based on $\Theta^{(i)}$ and $P^{(i)}$. Such a controller is known to be near-optimal when the parameter estimates $\Theta^{(i)}$ are sufficiently close to the true parameters~\cite{mania2019certainty}. In Appendix~\ref{app:lqr},
We provide a brief explanation about how the loss function~\eqref{eq:lq-tr-obj} effectively incapsulates the task-relevant information.

\section{Empirical Evaluation}
In this section, we empirically validate the effectiveness of using the task-relevant loss function and test the proposed method in a complex control problem.

\subsection{Meta-RL}
\begin{figure}[t]
\centering
\begin{subfigure}{0.872\textwidth}
\includegraphics[width=\textwidth]{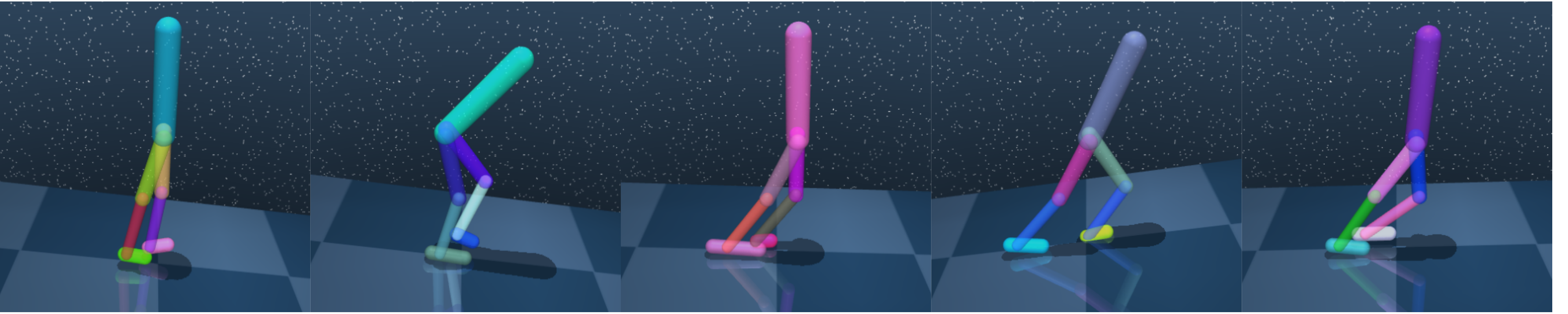}
\caption{}
\label{fig:walker}
\end{subfigure} \\
\begin{subfigure}{0.46\textwidth}%
\includegraphics[width=\textwidth]{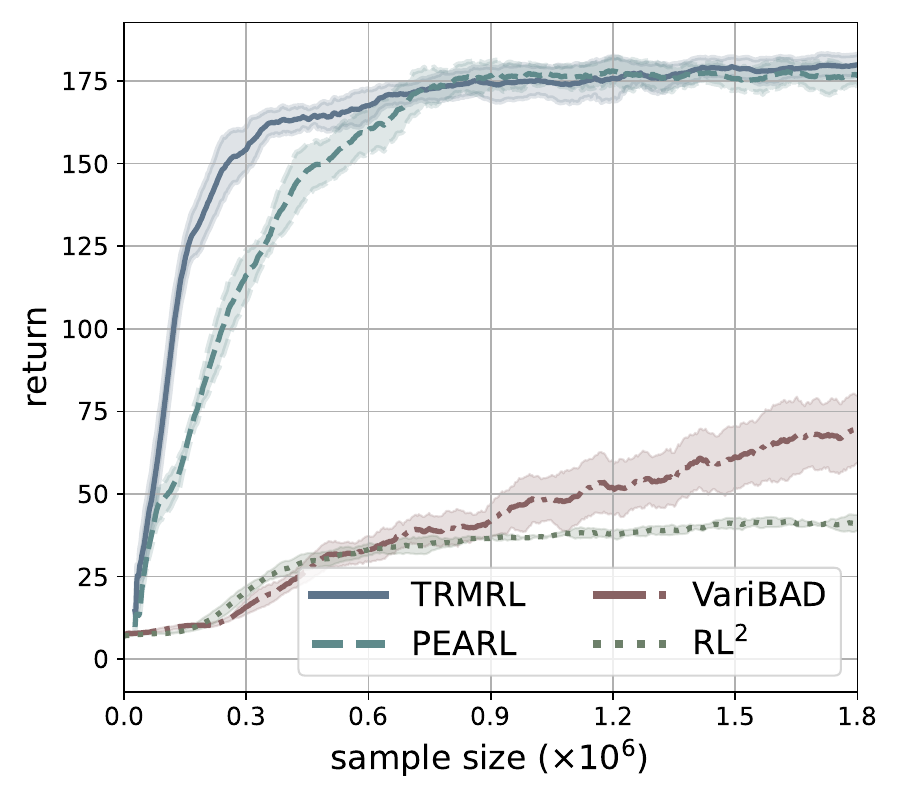}
\caption{}
\label{fig:res-baseline}
\end{subfigure}
\begin{subfigure}{0.46\textwidth}%
\includegraphics[width=\textwidth]{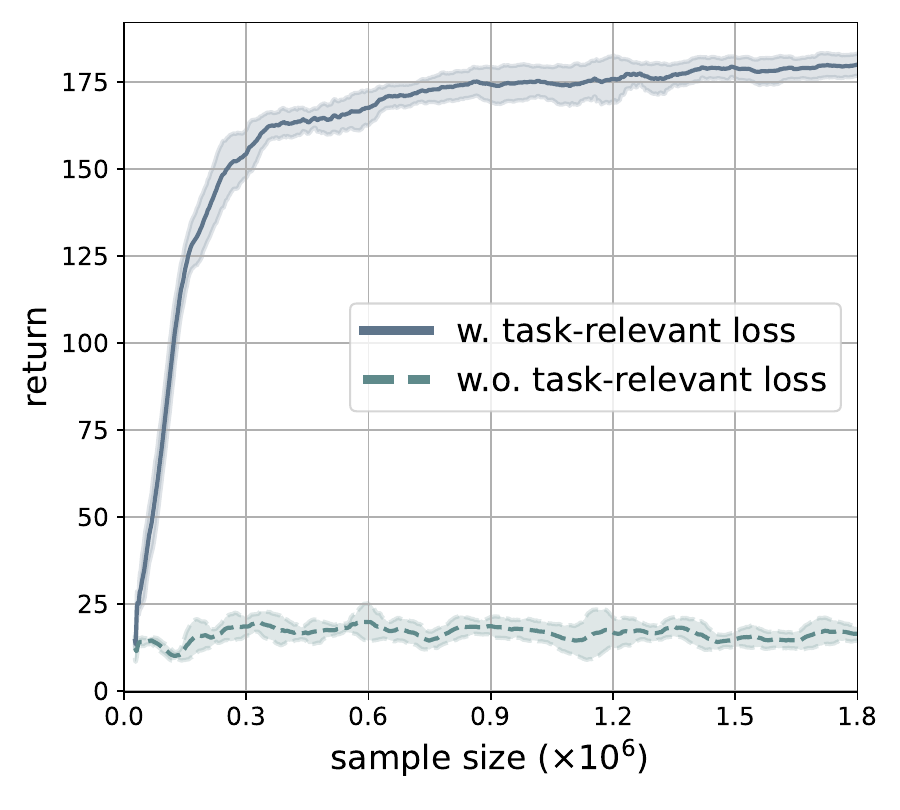}
\caption{}
\label{fig:res-task-rel}
\end{subfigure}
\caption{\textit{(a)} Illustration of the bipedal walker control problem with varying physical parameters. \textit{(b)} Comparison of the proposed method and the baseline methods. \textit{(c)} Demonstration of the effectiveness of the task-relevant loss. All methods are evaluated for 200 steps in each test task, and their returns are plotted in the $y$-axis. As the reward function of each task is designed to take values in $[0, 1]$, an upper bound of the return per episode is 200.}
\label{fig:res}
\end{figure}

We first evaluate the presented method in a robotic control problem, where the physical properties of the system vary across tasks. Specifically, we consider the bipedal walker modeled in \textsc{DeepMind Control Suite}~\cite{tunyasuvunakool2020} and simulated by \textsc{MuJoCo} engine~\cite{todorov2012mujoco}. The robot has the state space $\mathcal{S} \subseteq \mathbb{R}^{26}$ and the action space $\mathcal{A} = [-1, 1]^6$, and its objective is to walk along a sloped terrain at a steady speed of 1\textit{m/s}. However, the physical parameters governing the dynamics of the robot, such as its \textit{mass, inertia, foot length, joint damping coefficients, and link friction coefficients}, and the \textit{slope angle of the terrain}, differ for each task as shown in Figure~\ref{fig:walker}.\footnote{In particular, the presence of the slope makes the reward functions of each task differ from each other, as the reward is determined by measuring the speed along the terrain rather than the speed along the $x$-axis.} These parameters are selected from the predetermined probability distributions, where we sample 120 distinct training tasks.  Three state-of-the-art meta-RL algorithms, RL$^2$~\cite{duan2016rl}, PEARL~\cite{rakelly2019efficient}, and VariBAD~\cite{zintgraf2019varibad}, are used for performance comparison. 

Figure~\ref{fig:res-baseline} compares the performance of the proposed methods and the baseline algorithms. As a model-based method, TRMRL enjoys sample efficiency, exhibiting a faster learning speed than PEARL and better sample efficiency than the on-policy methods VariBAD and RL$^2$ in terms of sample efficiency. Specifically, the quality of TRMRL policies dramatically improves until 400,000 samples are collected, which amounts to less than 4 hours of operating the robot in 40 \textit{Hz}. In addition to its learning efficiency, TRMRL is comparable to PEARL concerning the quality of the learned controllers. In Figure~\ref{fig:res-task-rel}, we also illustrate the advantage of using the task-relevant loss by comparing our method and a method that uses a naive least square loss. Surprisingly, using a least square loss does not improve the quality of the learned policies.

\subsection{Online LQR}
\begin{figure}[t]
\centering
\begin{subfigure}{0.46\textwidth}%
\includegraphics[width=\textwidth]{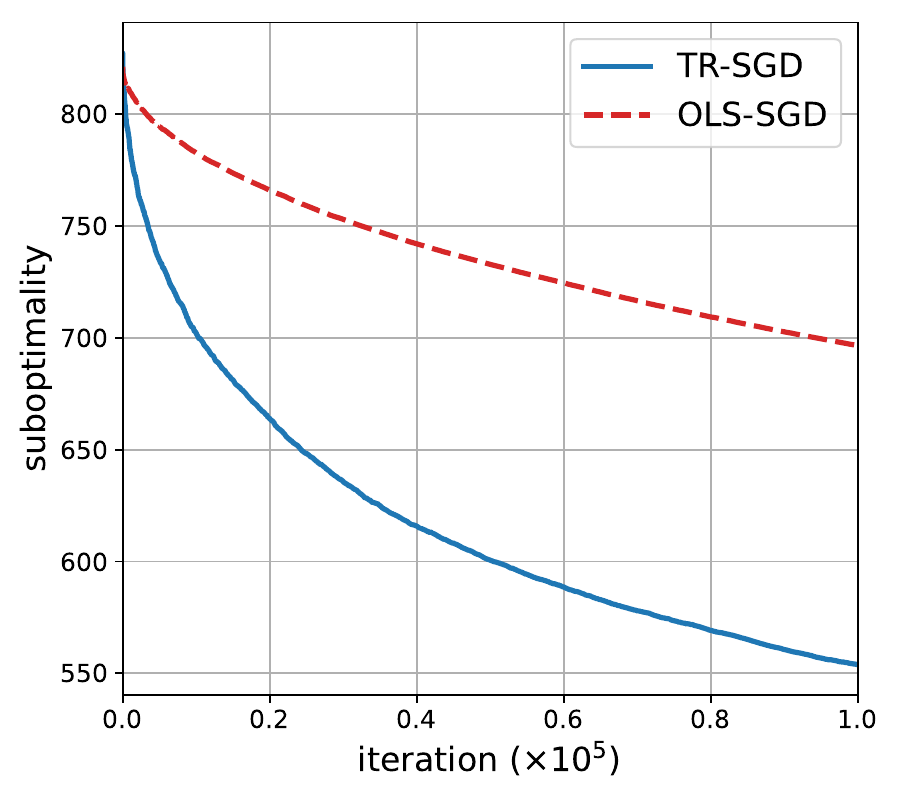}
\caption{}
\label{fig:lq-value}
\end{subfigure}
\begin{subfigure}{0.46\textwidth}%
\includegraphics[width=\textwidth]{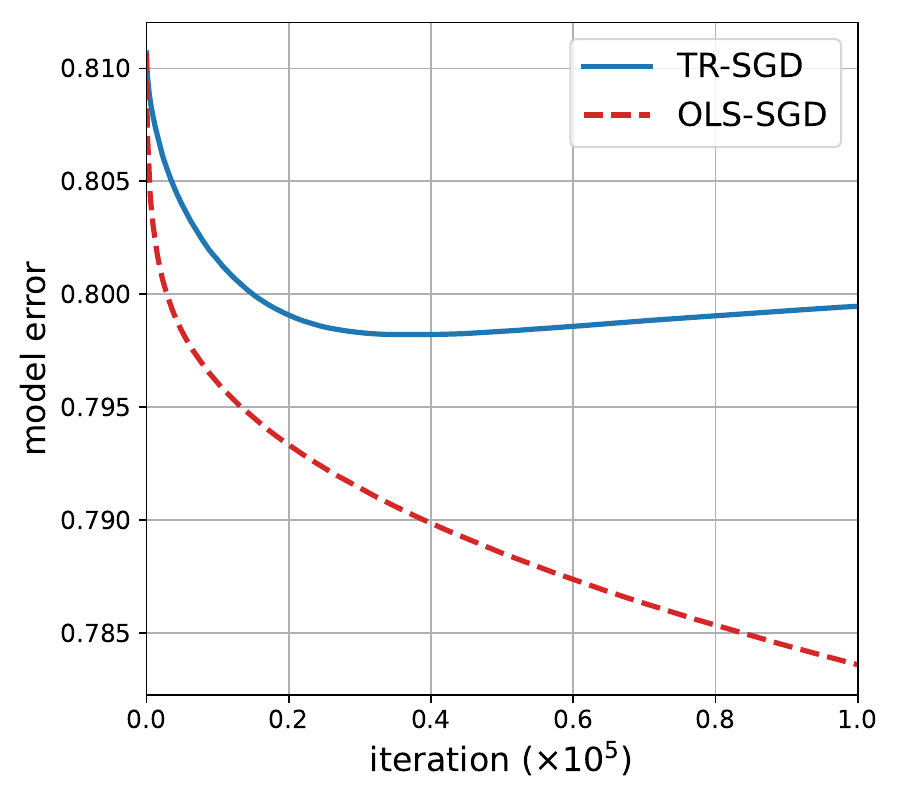}
\caption{}
\label{fig:lq-model}
\end{subfigure}
\caption{Illustration of the learning curves in a 20-dimensional online LQR task. The learning curve of the method with the task-relevant loss is denoted by blue solid lines, while the method using the OLS loss is represented as red dashed lines. \textit{(a)} exhibits the suboptimality of the policies, and \textit{(b)} shows the model errors measured in $2$-norm.}
\label{fig:lq}
\end{figure}
We demonstrate the advantage of considering task-relevant loss function by empirically investigating the behavior of \textbf{\texttt{TR-SGD}} and \textbf{\texttt{OLS-SGD}} in the LQR problem. To do so, we consider an LQR problem with $n=m=20$, where $A$ and $B$ are generated by randomly sampling the entries:
\begin{equation*}
a_{ij} \sim \textbf{UNIFORM}\left[0, 0.3\right], \quad b_{ij} \sim \textbf{UNIFORM}\left[0, 0.1 \right],
\end{equation*}
and are checked if $(A, B)$ is controllable and $A$ is unstable.
The initial candidate $\Theta^{(i)}$ is obtained by distorting the true parameter $\Theta$ through rank-one perturbation to prevent the differences $\|A - A^{(0)} \|_2$ and $\|B - B^{(0)} \|_2$ from being unmanageably large. Furthermore, the trajectories of length 10 are collected and used to update the parameters of both \textbf{\texttt{TR-SGD}} and \textbf{\texttt{OLS-SGD}}. For these methods, the regularization coefficient is set to $\lambda = 10^{-5}$, and both $\alpha_{\text{OLS}}^{(i)}$ and $\alpha_{\text{TR}}^{(i)}$ are chosen to be proportional to $(1 + i)^{-(1/2 + \varepsilon)}$ for some small $\varepsilon > 0$ in order to satisfy the  Robbins-Monro condition~\cite{robbins1951stochastic}.

Figure~\ref{fig:lq} compares the performances of the \textbf{\texttt{TR-SGD}} and \textbf{\texttt{OLS-SGD}}. In Figure~\ref{fig:lq-value}, the benefit of using the task-relevant loss is clearly confirmed by comparing the value function $P^{\pi^{(i)}}$ of the obtained policies with the optimal value function $P^\star$. Specifically, the value function of the policy $\pi^{(i)}(x) = K^{(i)}x$ is evaluated by solving the following \textit{discrete-time Lyapunov equation}
\begin{equation*}
P = A_{\text{cl}}^\top P A_{\text{cl}} + \mathbf{Q} +  {K^{(i)}}^\top \mathbf{R} K^{(i)},
\end{equation*}
where $A_{\text{cl}} \coloneqq A + B K^{(i)}$ is a closed loop matrix of the true system induced by $\pi^{(i)}$. Then, the value function error $\Vert P^{\pi^{(i)}} - P^\star \Vert_2$ represents the magnitude of the suboptimality incurred by using $\pi^{(i)}$:
\begin{equation*}
\Vert P^{\pi^{(i)}} - P^\star \Vert_2 = \lambda_{\max}(P^\star - P^{\pi^{(i)}}) = \sup_{\| x_0 \|_2 \leq 1} | V^\star(x_0) - V^{\pi^{(i)}}(x_0) |.
\end{equation*}
Therefore, Figure~\ref{fig:lq-value} indicates that using the task-relevant loss function~
\eqref{eq:lq-tr-gd} significantly accelerates the improvement of the resulting value function. This is because the loss~\eqref{eq:lq-tr-obj} directly incorporates the primary objective of the LQR, namely its value function, into the learning procedure. Such an effect is further substantiated in Figure~\ref{fig:lq-model} where the model error $\Vert \Theta^{(i)} - \Theta\Vert_2$ of \textbf{\texttt{OLS-SGD}} matches or surpasses that of the task-relevant method. This confirms that our method focuses on identifying the part of the dynamics that is crucial for obtaining an efficient controller, rather than only predicting the state transitions correctly.

\section{Conclusion}
To alleviate the issue of sample inefficiency in meta-RL,
we proposed TRMRL, a model-based method that
uses a carefully designed task-relevant loss function for both the task inference module and the system or environment model.
 This was inspired by the policy suboptimality bound that indicates the significance of measuring the value function discrepancy for learning the environmental model. The efficiency of the strategy was demonstrated by empirically evaluating it in  the bipedal walker control problem under large environmental changes and the LQR problem. Among numerous promising directions for future research are  extending our method to directly integrate raw sensor data and applying the method to real-world robotic systems.

\appendix
\section{Proof of Theorem~\ref{thm:main}}\label{app:proof}
\begin{proof}
Throughout the proof, we suppress the subscript $\infty$ of $\Vert \cdot\Vert_\infty$ for simplicity.
To begin with, we introduce the following Bellman operators:
\begin{align*}
(F_M^\pi V)(s) &\coloneqq \int_{a \in \mathcal{A}} \pi(da|s) \Big( R(s, a) + \gamma TV(s, a) \Big), \quad s \in \mathcal{S}, \\
(F_M V)(s) &\coloneqq \sup_{a} \Big( R(s, a) + \gamma TV(s, a) \Big), \quad s \in \mathcal{S}.
\end{align*}
Then, we use the triangle inequality to decompose the suboptimality gap as follows:
\begin{align*}
\Vert V_M^\star - V_M^{\pi_{\latent}} \Vert &\leq \underbrace{\Vert V_{M_{\latent}}^{\pi_{\latent}} - V_M^{\pi_{\latent}} \Vert + \Vert V_M^\star - V_{M_{\latent}}^\star \Vert}_{\text{task mismatch error}}  + \underbrace{\Vert V_{M_{\latent}}^\star - V_{M_{\latent}}^{\pi_{\latent}} \Vert}_{\text{planning error}}. 
\end{align*}
First of all, the planning error is bounded by applying Proposition~6.1 of~\cite{bertsekas1996neuro}:
\begin{equation}\label{pf:block1}\tag{A.1}
\Vert V_{M_{\latent}}^\star - V_{M_{\latent}}^{\pi_{\latent}} \Vert \leq \frac{2\gamma}{1-\gamma}\underbrace{\Vert V^\star_{M_{\latent}} - V_{\latent} \Vert}_{\leqslant \varepsilon} \leq \frac{2\gamma\varepsilon}{1 - \gamma},
\end{equation}
where we use the assumption that the approximate value function $V_{\latent}$ differs from the optimal one $V^\star_{M_{\latent}}$ by at most $\varepsilon$ in terms of $\ell^\infty$-norm.
Furthermore, the first term of the task mismatch error, which measures the variation of the policy's value across the tasks, is bounded as follows:
\begin{align*}
\Vert V_M^{\pi_{\latent}} - V_{M_{\latent}}^{\pi_{\latent}} \Vert &= \Vert F^{\pi_{\latent}}_M(V_M^{\pi_{\latent}}) - F^{\pi_{\latent}}_{M_{\latent}}(V^{\pi_{\latent}}_{M_{\latent}}) \Vert \\
&\leq \Vert F_M^{\pi_{\latent}}(V^{\pi_{\latent}}_{M_{\latent}}) - F_{M_{\latent}}^{\pi_{\latent}}(V^{\pi_{\latent}}_{M_{\latent}}) \Vert + \Vert F_{M}^{\pi_{\latent}}(V^{\pi_{\latent}}_M) - F_{M}^{\pi_{\latent}}(V^{\pi_{\latent}}_{M_{\latent}}) \Vert \\
&\leq \Vert F_M^{\pi_{\latent}}(V^{\pi_{\latent}}_{M_{\latent}}) - F_{M_{\latent}}^{\pi_{\latent}}(V^{\pi_{\latent}}_{M_{\latent}}) \Vert + \gamma \Vert V_M^{\pi_{\latent}} - V_{M_{\latent}}^{\pi_{\latent}} \Vert, \\
\Longrightarrow \quad \Vert V_M^{\pi_{\latent}} - V_{M_{\latent}}^{\pi_{\latent}} \Vert &\leq \frac{1}{1 - \gamma} \underbrace{\Vert F_M^{\pi_{\latent}}(V^{\pi_{\latent}}_{M_{\latent}}) - F_{M_{\latent}}^{\pi_{\latent}}(V^{\pi_{\latent}}_{M_{\latent}}) \Vert}_{\coloneqq (A)} ,
\end{align*}
since $V_M^{\pi_{\latent}}$ and $V_{M_{\latent}}^{\pi_{\latent}}$ are the fixed points of $F^{\pi_{\latent}}_M$ and $F^{\pi_{\latent}}_{M_{\latent}}$, respectively, and $F^{\pi_{\latent}}_M$ is a $\gamma$-contraction. However, note that
\begin{align*}
(A) &= \sup_{s} \left| \int_{\mathcal{A}}  \left( R(s, a) - R_{\latent}(s, a) + \gamma(T - T_{\latent})V^{\pi_{\latent}}_{M_{\latent}}(s, a)  \right) \pi_{\latent}(da|s)   \right| \\
&\leq \sup_{s}  \int_{\mathcal{A}}  \left| R(s, a) - R_{\latent}(s, a) + \gamma(T - T_{\latent})V^{\pi_{\latent}}_{M_{\latent}}(s, a) \right| \pi_{\latent}(da|s) \\
&\leq \sup_{s, a} \left| R(s, a) - R_{\latent}(s, a) + \gamma(T - T_{\latent})V^{\pi_{\latent}}_{M_{\latent}}(s, a) \right| \\
&= \left\Vert (R - R_{\latent}) + \gamma\left(T - T_{\latent} \right) V^{\pi_{\latent}}_{M_{\latent}} \right\Vert,
\end{align*}
where we use $\int_{\mathcal{A}} \pi_{\latent}(da|s) = 1$ for all $s\in \mathcal{S}$.
Thus, we have
\begin{align*}
\Vert V_M^{\pi_{\latent}} - V_{M_{\latent}}^{\pi_{\latent}} \Vert \leq \frac{1}{1 - \gamma} \left\Vert (R - R_{\latent}) + \gamma\left(T - T_{\latent} \right) V^{\pi_{\latent}}_{M_{\latent}} \right\Vert.
\end{align*}
Similarly, we bound the second term of the task mismatch error concerning the dependency of the optimal value function with respect to the choice of a task as follows:
\begin{align*}
\Vert V_M^\star - V_{M_{\latent}}^\star \Vert &= \Vert F_M (V^\star_M) - F_{M_{\latent}} (V^\star_{M_{\latent}})  \Vert \\
&\leq \Vert F_M (V^\star_M) - F_M(V^\star_{M_{\latent}}) \Vert + \Vert F_M(V^\star_{M_{\latent}}) - F_{M_{\latent}} (V^\star_{M_{\latent}})  \Vert \\
&\leq \gamma \Vert V_M^\star - V_{M_{\latent}}^\star \Vert + \Vert F_M(V^\star_{M_{\latent}}) - F_{M_{\latent}} (V^\star_{M_{\latent}})  \Vert, \\
\Longrightarrow \quad \Vert V_M^\star - V_{M_{\latent}}^\star \Vert &\leq \frac{1}{1 - \gamma} \underbrace{\Vert F_M(V^\star_{M_{\latent}}) - F_{M_{\latent}} (V^\star_{M_{\latent}})  \Vert}_{\coloneqq (B)}.
\end{align*}
Then, we deduce
\begin{align*}
(B) &= \sup_{s} \left| \sup_a \left( R(s, a) + \gamma T V^\star_{M_{\latent}} (s, a) \right) - \sup_a  \left( R_{\latent}(s, a) + \gamma T_{\latent} V^\star_{M_{\latent}} (s, a) \right)  \right| \\
&\leq \sup_s \left| \sup_{a} \left( R(s, a) - R_{\latent}(s, a) + \gamma (T - T_{\latent}) V^\star_{M_{\latent}}(s, a)      \right) \right| \\
&\leq \sup_{s, a} \left|  R(s, a) - R_{\latent}(s, a) + \gamma(T - T_{\latent})V^\star_{M_{\latent}}(s, a)  \right| \\
&= \left\Vert (R - R_{\latent}) + \gamma\left(T - T_{\latent} \right) V^\star_{M_{\latent}} \right\Vert
\end{align*}
from which we obtain
\begin{equation*}
\Vert V_M^\star - V_{M_{\latent}}^\star \Vert \leq \frac{1}{1 - \gamma} \left\Vert (R - R_{\latent}) + \gamma\left(T - T_{\latent} \right) V^\star_{M_{\latent}} \right\Vert.
\end{equation*}
However, as $\Vert V^\star_{M_{\latent}} - V_{\latent} \Vert \leq \varepsilon$ we have
\begin{align}\label{pf:block2}
\left\Vert (R - R_{\latent}) + \gamma\left(T - T_{\latent} \right) V^\star_{M_{\latent}} \right\Vert &\leq \left\Vert (R - R_{\latent}) + \gamma\left(T - T_{\latent} \right) V_{\latent} \right\Vert \nonumber\\
&\quad + \left\Vert \left(T - T_{\latent} \right) (V^\star_{M_{\latent}} - V_{\latent}) \right\Vert \nonumber \\ 
&\leq \left\Vert (R - R_{\latent}) + \gamma\left(T - T_{\latent} \right) V_{\latent} \right\Vert + 2\varepsilon, \nonumber\\
\Longrightarrow\quad \Vert V_M^\star - V_{M_{\latent}}^\star \Vert &\leq \frac{1}{1 - \gamma}\left\Vert (R - R_{\latent}) + \gamma\left(T - T_{\latent} \right) V_{\latent} \right\Vert + \frac{2\varepsilon}{1 - \gamma}. \tag{A.2}
\end{align}
Furthermore, an analogous argument leads to
\begin{align}\label{pf:block3}
\left\Vert (R - R_{\latent}) + \gamma\left(T - T_{\latent} \right) V^{\pi_{\latent}}_{M_{\latent}} \right\Vert 
\leq \left\Vert (R - R_{\latent}) + \gamma\left(T - T_{\latent} \right) V_{\latent} \right\Vert + 2 \frac{1 + \gamma}{1 - \gamma}\varepsilon, \nonumber\\
\Longrightarrow\quad \Vert V_M^{\pi_{\latent}} - V_{M_{\latent}}^{\pi_{\latent}} \Vert \leq \frac{1}{1 - \gamma}\left\Vert (R - R_{\latent}) + \gamma\left(T - T_{\latent} \right) V_{\latent} \right\Vert + \frac{2(1 + \gamma)\varepsilon}{(1 - \gamma)^2} \tag{A.3}
\end{align}
because   $\Vert V^{\pi_{\latent}}_{M_{\latent}} - V_{\latent} \Vert_\infty \leq \frac{1 + \gamma}{1 - \gamma} \varepsilon$. (This can be shown by using a technique similar to the one used in Proposition 6.1 of~\cite{bertsekas1996neuro}.)
Finally, combining~\eqref{pf:block1}, \eqref{pf:block2}, and~\eqref{pf:block3}, the result follows.
\end{proof}

\section{Intuition from Case Analysis: Online LQR}\label{app:lqr}
In this section, we aims to understand provide a high-level explanation about the reason for the success of using the task-relevant loss function in online LQR by investigating \textbf{\texttt{TR-SGD}}. To do so, we first note that its baseline method \textbf{\texttt{OLS-SGD}} may be viewed as solving the following stochastic optimization problem via SGD:
\begin{equation}\label{eq:opt-ols}
\min_{\Theta} J_{\mathrm{OLS}}(\Theta) \coloneqq \mathbb{E}_{\tau} \left[ J_{\mathrm{OLS}} (\Theta|\tau)\right], \quad \Theta\in \mathbb{R}^{n \times (n+m)}.
\end{equation}
where the expectation is taken over the distribution of the trajectories generated by executing the random controller.
Since $J_{\mathrm{OLS}} (\Theta|\tau)$ is convex \textit{w.r.t.} $\Theta$, \textbf{\texttt{OLS-SGD}} almost surely converges to the unique minimum $\Theta^\star$ of $J_{\mathrm{OLS}}$ if the objective function is not regularized, {i.e.}, $\lambda = 0$, and the stepsizes $\alpha^{(i)}_{\mathrm{OLS}}$ satisfy Robbins-Monro condition. The point is that the minimizer of $J_{\mathrm{OLS}}$ is unique; $J_{\mathrm{OLS}}$ is optimized only when the true parameter $\Theta^\star$ is completely identified. Thus, the convergence toward the minimizer may be slow depending on how the initial iterate $\Theta^{(0)}$ is chosen.
On the other hand, consider the following minimization problem associated with  \textbf{\texttt{OLS-SGD}}:
\begin{equation}\label{eq:opt-tr}
\min_{\Theta} J_{\mathrm{TR}}(\Theta|P) \coloneqq \mathbb{E}_{\tau} \left[ J_{\mathrm{TR}} (\Theta|\tau, P)\right], \quad \Theta\in \mathbb{R}^{n \times (n+m)},
\end{equation}
where $P$ is a fixed positive-definite matrix. Then, it is clear that $J_{\mathrm{TR}}(\cdot|P)$ achieves its minimum 0 at $\Theta$ if $\Theta$ satisfies
\begin{equation}\label{eq:lq-minimizer}
\Theta^\top P \Theta = {\Theta^\star}^\top P \Theta^\star.
\end{equation}
Therefore, the set of the minimizers of $J_{\mathrm{TR}}(\cdot | P)$ is larger than the singleton $\{ \Theta^\star \}$. Indeed, if we choose an arbitrary matrix $V$ from the orthogonal group $O(n)$ and let
\begin{equation*}
\Theta = L^{-1} V L \Theta^\star,
\end{equation*}
where $L \coloneqq P^{1/2}$,
then it is straightforward to check that $\Theta$ satisfies the identity~\eqref{eq:lq-minimizer}. Therefore, the set $\mathscr{O} \coloneqq \{  L^{-1} V L \Theta^\star: V\in O(n) \}$ is contained in the set of minimizers of $J_{\mathrm{TR}}(\cdot|P)$. Since $\mathscr{O}$ is an orbit $G \cdot \Theta^\star$ under the action of the group
\begin{equation*}
G \coloneqq \{L^{-1} V L: V \in O(n)  \},
\end{equation*}
which is isomorphic to $O(n)$ and $\dim O(n) = \frac{n(n-1)}{2}$, the set of minimizers contains a manifold of dimension $\frac{n(n-1)}{2}$. Therefore, it suffices for the iterates $\Theta^{(i)}$ of~\eqref{eq:opt-tr} to approach toward $\mathscr{O}$ rather than the true parameter $\Theta^\star$, which may accelerate the convergence of SGD. Furthermore, all the parameters $\Theta \in \mathscr{O}$ are indistinguishable from the true parameter $\Theta^\star$ if one employs the greedy policy
\begin{equation*}
\pi(x) = \argmin_{u} \left(r(x, u) - \gamma [x^\top\; u^\top] \Theta^\top P \Theta \begin{bmatrix} x \\ u \end{bmatrix} \right), \quad x \in \mathbb{R}^n,
\end{equation*}
which becomes optimal if $P = P^\star$. The arguments indicate that our loss function benefits from considering the `symmetry' of the problem; since the optimal value function is given as $V(x) = -x^\top P^\star x$, all state $x$ with the same norm $\Vert x \Vert_P = \sqrt{\langle x, Px \rangle}$ are equivalent, which eliminates the need for exactly identifying the state. We rather require the model to ensure that the predicted state is located on the same ellipsoid $\mathscr{E}_\delta(P) \coloneqq \{y: y^\top P y = \delta \}$ with the true state. This amounts to utilizing the symmetry of the ellipsoid $\mathscr{E}_\delta(P)$ which is explicitly represented as $G$.
While \textbf{\texttt{TR-SGD}} uses $P^{(i)}$ instead of the unknown $P^\star$, the argument provides an insight into why the proposed method outperforms the naive \textbf{\texttt{OLS-SGD}} from the optimization perspective. Even though the above argument is only valid for LQR, we believe that our loss function works similarly by recognizing the symmetry and `compressing' the environment in general non-linear problems.

\bibliographystyle{IEEEtran}
\bibliography{l4dc2024}

\end{document}